\documentclass[11pt,a4paper]{article}
\usepackage[hyperref]{styles/acl2021}
\usepackage{times}
\usepackage{latexsym}
\renewcommand{\UrlFont}{\ttfamily\small}

\usepackage{microtype}

\usepackage{graphicx}
\usepackage{color}
\usepackage{amsmath}
\usepackage{multirow}
\usepackage{enumitem}

\usepackage{float} 

\aclfinalcopy 
\def\aclpaperid{3592} 

\newcommand{\reffig}[1]{Figure~\ref{#1}}
\newcommand{\reftbl}[1]{Table~\ref{#1}}
\newcommand{\refsec}[1]{Section~\ref{#1}}
\newcommand{\refapp}[1]{Appendix~\ref{#1}}

\newcommand{\reminder}[1]{}

\newcommand{\system}{PERO}

\newcommand{\systemfull}{Prompting with Examples in the Right Order}

\newcommand{\textformat}[2]{$\mathrm{Format}$({#1},{#2})}

\newcommand{\separator}{\textless Separator\textgreater}

\usepackage{array}
\newcolumntype{L}[1]{>{\raggedright\let\newline\\\arraybackslash\hspace{0pt}}m{#1}}
\newcolumntype{C}[1]{>{\centering\let\newline\\\arraybackslash\hspace{0pt}}m{#1}}
\newcolumntype{R}[1]{>{\raggedleft\let\newline\\\arraybackslash\hspace{0pt}}m{#1}}

\title{Reordering Examples Helps during Priming-based Few-Shot Learning}

\author{Sawan Kumar \\
  Indian Institute of Science, Bangalore \\
  \texttt{sawankumar@iisc.ac.in} \\\And
  Partha Talukdar \\
  Indian Institute of Science, Bangalore \\
  \texttt{ppt@iisc.ac.in} \\}

\date{}

\begin{document}
\maketitle
\begin{abstract}
The ability to learn from limited data, or few-shot learning, is a desirable and often critical requirement for NLP systems. While many existing methods do poorly at learning from a handful of examples, large pretrained language models have recently been shown to be efficient few-shot learners. One approach to few-shot learning, which does not require finetuning of model parameters, is to augment the language model's input with priming text which is typically constructed using task specific descriptions and examples. In this work, we further explore priming-based few-shot learning, with focus on using examples as prompts. We show that presenting examples in the right order is key for generalization. We introduce \system{} (\systemfull{}), where we formulate few-shot learning as search over the set of permutations of the training examples. We show that \system{} can learn to generalize efficiently using as few as 10 examples, in contrast to existing approaches. While the newline token is a natural choice for separating the examples in the prompt, we show that learning a new separator token can potentially provide further gains in performance. We demonstrate the effectiveness of the proposed method on the tasks of sentiment classification, natural language inference and fact retrieval. Finally, we analyze the learned prompts to reveal novel insights, including the idea that two training examples in the right order alone can provide competitive performance for sentiment classification and natural language inference.

\end{abstract}

\section{Introduction}
\label{sec:intro}

The ability to learn from a few examples, or few-shot learning, as generally understood to be possessed by humans, is a desirable property for Natural Language Processing (NLP) systems as well. It is critical in scenarios where collecting large amounts of data is expensive. It is also important to enable a personalized Artificial Intelligence (AI) experience, where a single user is expected to use an AI agent to perform a task demonstrated through a handful of examples.\footnote{See the Introduction section of \citet{brown2020language} for a discussion on further difficulties, relevant to the setting we consider in this work, regarding the  need of a large dataset for every new task.}

\begin{figure*} [thb]
  \centering
  \includegraphics[width=1.0\linewidth]{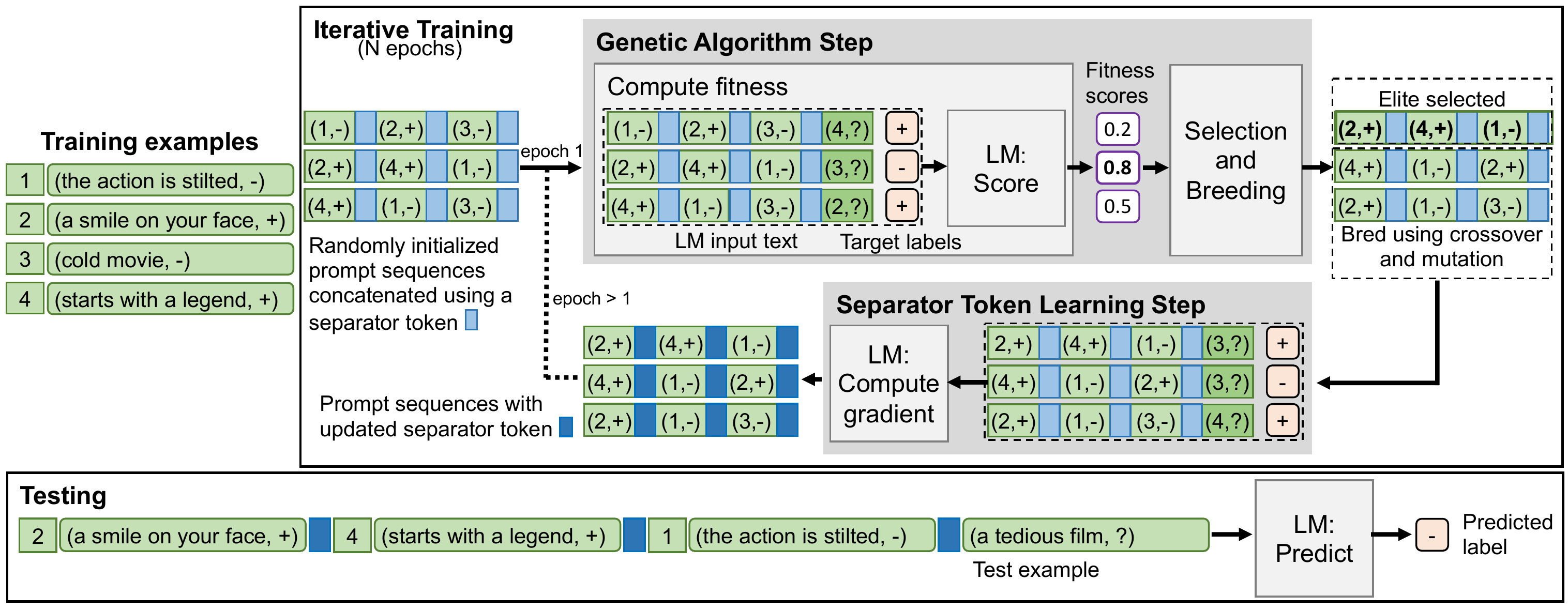}
  \caption{\label{fig:schematic} \textit{Overview of \system{}}: Given a set of training examples, \system{} searches over permutations of training examples using a genetic algorithm (\refsec{subsec:method_search}), and optionally also learns a separator token (\refsec{subsec:method_separator}) which is used to concatenate the training examples.  Briefly, starting from a set of randomly initialized permutations, the genetic algorithm step computes the fitness of each permutation for making predictions using a pretrained LM. These fitness scores are then used for selection and subsequent breeding of new permutations using biologically inspired operations of mutation and crossover. The separator token learning step uses the updated set of permutations and uses gradient updates to improve the separator token. The two steps are performed iteratively for a fixed number of epochs and the best permutation and separator token are selected using a validation set. Please see \refsec{sec:method} for details.}
\end{figure*}

Pretrained language models \citep{devlin-etal-2019-bert,liu2019roberta,raffel2019exploring} have recently been shown to be exceedingly good at several benchmark NLP tasks \citep{wang2018glue,wang2019superglue}. Traditionally the parameters of these language models have been finetuned on task specific datasets to achieve the aforementioned performance gains, often requiring large amounts of data. \citet{brown2020language} show that large pretrained language models (GPT3) are also efficient few-shot learners. Few-shot learning is achieved using task descriptions and labeled examples as prompts. Remarkably, with this priming-based approach and without needing any parameter updates, GPT3 often performs comparable to traditional finetuning-based supervised systems which use much larger datasets. One could argue that the task performance achieved in the priming-based approach measures what the pretrained language model has already learned. \citet{shin-etal-2020-autoprompt}, operating in the same setting, use automatically generated prompts to measure task specific knowledge in a pretrained language model.

In this work, we further explore priming-based few-shot learning, while focusing on using examples as prompts. The training objective for a language model is typically the prediction of a token given a context. There is no clear incentive to treat a sequence of sentences in the context as equal and conveying examples of a concept. As a result, one could expect certain order of examples when used as a prompt to be more favorable at providing task specific cues.

We propose \system{}\footnote{\system{} source code is available at \\
\UrlFont{https://github.com/SawanKumar28/pero}} (\systemfull{}), where we formulate the problem of few-shot learning as search over permutations of training examples. We find that choosing the right permutation is key to getting good task performance. In \system{}, we use a genetic algorithm \citep{mitchell1998introduction} to search over possible permutations of training examples. The selected examples are used for prompting publicly available pretrained language models \citep{devlin-etal-2019-bert,liu2019roberta}. We find that with as few as 10 examples, \system{} can learn to generalize efficiently, in contrast to existing approaches. When concatenating examples to use as a prompt, the newline token is a natural choice as a separator token. We show that using a learned separator token can potentially provide further gains in performance. We evaluate the performance of \system{} on the tasks of sentiment analysis, Natural Language Inference (NLI) and fact retrieval. 

Finally, our analysis of the learned prompts (\refsec{subsec:expts_analyze}) leads to novel insights about few-shot learning using textual prompts. For instance, using only two examples, repeated and ordered using a learned label pattern, can provide performance comparable to and even exceeding the performance of existing few-shot baselines which use thousands of examples.

In summary, we make the following contributions:
\begin{enumerate} 
	\item We propose \system{}, where we formulate the problem of few-shot learning as search over permutations of training examples, and optionally a separator token. As we don't update the parameters of the underlying language model, \system{} serves as a probe for measuring task specific knowledge in pretrained language models.
	\item We demonstrate the effectiveness of \system{} over a recent baseline on the tasks of sentiment analysis, NLI and fact retrieval.
	\item We analyze the learned prompts and provide novel insights about textual prompts that can lead to good task performance in the low-data regime. In particular, we provide an effective recipe for one-shot learning.
\end{enumerate}

We have released the source code of \system{} to aid reproducibility of the results.  

\begin{table*}[tbh]
\small
\centering
\begin{tabular}{|l|l|l|}
\hline
Task &
  Template &
  Examples \\ \hline
\begin{tabular}[c]{@{}l@{}}Sentiment \\ Classification\end{tabular} &
  {[}Example Text{]} Answer: {[}Label Text{]} &
  \begin{tabular}[c]{@{}l@{}}(1) goes to absurd length Answer: false\\ (2) A girl in white is dancing Answer: true\end{tabular} \\ \hline
NLI &
  \begin{tabular}[c]{@{}l@{}}``{[}Premise Text{]}" implies ``{[}Hypothesis text{]}"\\ Answer: {[}Label Text{]}\end{tabular} &
  \begin{tabular}[c]{@{}l@{}}(1) ``Men are sawing logs" implies ``Men are \\   cutting wood" Answer: true\\ (2) ``There is no girl in white dancing" implies\\   ``A girl in white is dancing" Answer: false\end{tabular} \\ \hline
Fact retrieval &
  \begin{tabular}[c]{@{}l@{}}{[}Subj{]} is located in {[}Obj{]}\\ {[}Subj{]} is a subclass of {[}Obj{]}\end{tabular} &
  \begin{tabular}[c]{@{}l@{}}(1) Directors Lounge is located in Berlin\\ (2) gingerbread is a subclass of cookie\end{tabular} \\ \hline
\end{tabular}
\caption{\label{table:tbl_formatting} \textit{Formatting used to create textual inputs} for the tasks considered in this work. For sentiment classification, positive and negative sentiments correspond to the label text of \textit{true} and \textit{false} respectively. For NLI, entailment and contradiction labels correspond to the label text of true and false respectively.}
\end{table*}

\section{Related Work}
\label{sec:related}

\textbf{Pretrained Language Models} using a transformer architecture \citep{vaswani2017attention} on large unsupervised corpora have recently been found to be efficient at learning downstream tasks, providing significant gains over existing standalone supervised systems, on a variety of NLP tasks \citep{wang2018glue,wang2019superglue}. There have been two major approaches to learning language models: causal language models (CLM) and masked language models (MLM). CLMs \citep{radford2018improving,radford2019language,brown2020language} are typically trained by requiring a language model to predict the next token given a textual context. Masked language models \citep{devlin-etal-2019-bert,liu2019roberta} on the other hand are trained  by masking out a certain number of tokens in a textual context and requiring the language model to predict the masked out tokens. Typically, the parameters of the language model are then finetuned using task-specific training examples. For our experiments, we leverage publicly available pretrained masked language models \citep{devlin-etal-2019-bert,liu2019roberta}.

\textbf{Few-shot learning using language models} is a desirable and perhaps even an expected property of large pretrained language models, given the large amounts of data they are typically trained with. \citet{brown2020language} show that scaling up language models leads to improved few-shot learning, with their best model, GPT3, being able to achieve performance comparable to existing supervised systems, while using much fewer examples. Zero-shot and few-shot learning are achieved without needing parameter updates to the model but instead by prompting the language model with task specific description and task specific examples. In this work, we study the impact of the order in which examples are presented in a prompt and show that searching over them can lead to significant gains in few-shot performance, without needing updates to the model parameters.\footnote{Note that the scope of this work is distinct from meta-learning approaches \citep{hospedales2020meta}, where the goal is improve the learning algorithm using several learning episodes. In contrast, we only assume a pretrained language model and a few examples of the concept we are interested in.}

Measuring task performance of language models without any parameter updates can be seen as a measure of the knowledge (either descriptive, or procedural) that is already contained in the pretrained language model.

\textbf{Probing knowledge contained in language models} has been of interest, given the success of these models. Probing methods rely on creating cloze-style manual prompts \citep{petroni-etal-2019-language}, or mining efficient natural language prompts \citep{jiang-etal-2020-know}. \citet{shin-etal-2020-autoprompt} rely on training examples to learn trigger tokens which when used as a prompt demonstrate the ability of language models to do sentiment analysis and NLI along with knowledge based completion, without needing any parameter updates. The learned trigger tokens however aren't very meaningful leading to difficulty in interpreting these results. In this work, we instead focus on using natural language training examples as prompts. While being more interpretable, the prompts used in this work lead to significant gains in performance in the low-data regime.

\section{Background: Genetic Algorithm}
\label{subsec:background_genetic}
A genetic algorithm \citep{mitchell1998introduction} is a search heuristic inspired by the biological process of natural selection. Briefly, it evolves a population of candidate solutions towards increasing fitness to an objective through biologically inspired operations such as selection, crossover and mutation. We now describe the key terminology:
\begin{description} 
	\item[Individual] A single candidate solution, $c$, usually represented through a binary code but extensible to other types of codes. Generally, we will let $c$ be denoted by the sequence of $k$ integers $c= ( s^1s^2...s^k )$.
	\item[Population] A set of individuals of size $N_P$, $P=\{c_i, i \in [N_P]\}$.
	\item[Fitness] The measure of goodness for an individual for the task, $F(c_i)$.
	\item[Selection] An operator to select fit individuals in a population which will be used to generate new individuals, through crossover and mutation. Better fitness leads to higher likelihood of selection.
	\item[Crossover] An operator which typically takes two individuals $c_1$ and $c_2$ as inputs to produce new individuals $d_1$ and $d_2$, by combining sub-sequences from the two inputs. For example, consider two input sequences:
\begin{align*}
	c_1&=(c_1(1)c_1(2)c_1(3)c_1(4)) \\
	c_2&=(c_2(1)c_2(2)c_2(3)c_2(4))
\end{align*}	
A single point crossover after the second position would lead to the individuals:
\begin{align*}
	d_1&=(c_1(1)c_1(2)c_2(3)c_2(4)) \\
	d_2&=(c_2(1)c_2(2)c_1(3)c_1(4))
\end{align*}	
	\item[Mutation] An operator which randomly flips some elements in an input sequence. For example, with input $c=(c(1)c(2)c(3)c(4))$, a typical mutation operation would lead to the output $d=(c(1)c(2)c'(3)c(4))$, where $c'(3) \neq c(3)$. Usually, each position is randomly altered with a mutation probability $p_m$.
\end{description}

We now present the sketch of a typical genetic algorithm:
\begin{enumerate} 
	\item Initialize a set of individuals to form a population $P=\{c_i, i \in [N_P]\}$. Repeat the following steps for $N_{\mathrm{epochs}}$ iterations.
	\item Compute fitness of each individual in the population, $F(c_i), i \in [N_P]$.
	\item Using the computed fitness, select individuals which will be used to breed the next generation.
	\item With pairs of selected individuals, generate new individuals using the crossover operation.
	\item Mutate the generated individuals using the mutation operator, to create a new population $P'$.
	\item Set $P=P'$ and go to step 2.
\end{enumerate}

\section{\system{}: Proposed Method}
\label{sec:method}

The overall architecture employed in PERO is shown in \reffig{fig:schematic}. We introduce the notation in \refsec{subsec:method_notation}. We discuss how we employ a genetic algorithm to search over permutations of training examples in \refsec{subsec:method_search}. We then discuss how we augment the search heuristic to learn a task specific separator token in \refsec{subsec:method_separator}.

\subsection{Notation and Input Format}
\label{subsec:method_notation}

For both classification and knowledge base completion tasks, we denote a textual task input by $x$ and the gold label as $y$. We denote the pretrained masked language model with the operator $L$, which takes a sequence of input tokens to output a sequence of the same length containing token probabilities over the token vocabulary. With tokens $(t_1t_2...t_n)$, $L((t_1t_2...t_n))=(p_1p_2...p_n)$, where $p_i$ denotes a vector of probabilities over all tokens in the vocabulary.

For all our experiments, the input to the language model is formatted with exactly one mask token.\footnote{The mask token we use is the same as the one employed in pretraining.} For brevity, we denote by $L_{\mathrm{Mask}}$ the operator which outputs the token probability at the mask token position. 

The training data is denoted by the set of examples $(x_i, y_i), i \in [N_{\mathrm{train}}]$. We denote a permutation, or an ordered subset of size $k$ of the training data, by $c=(c(1)c(2)...c(k))$, where $c(j) \in [N_{\mathrm{train}}]$.

For all tasks, we create an input text sequence by concatenating $k$ examples using a permutation $c$ of training examples, along with a test example $x_\mathrm{test}$: ``\textformat{$x_{c(1)}$}{$y_{c(1)}$} \separator{} \textformat{$x_{c(2)}$}{$y_{c(2)}$} .. \separator{} \textformat{$x_{c(k)}$}{$y_{c(k)}$} \separator{} \textformat{$x_{test}$}{$\mathrm{mask}$}", where \textformat{}{} formats the example text and label for a task, and \separator{} is either the new line character, or is learned as described in \refsec{subsec:method_separator}. The formatting details are provided in \reftbl{table:tbl_formatting}. We attempt to use task agnostic formats and textual labels for classification tasks to the extent possible.

\subsection{Genetic Algorithm: Search over Permutations of Examples}
\label{subsec:method_search}
We employ a genetic algorithm for searching over permutations of training examples (see \refsec{subsec:background_genetic} for a brief introduction to genetic algorithms). We present the overall architecture in \reffig{fig:schematic}.

 Here, we detail how the various components and operators of a genetic algorithm are defined for searching over permutations of examples:

\begin{description} 
	\item[Individual] An individual is defined as a vector of unique training example indices  $c=(c(1)c(2)...c(k))$, where $c(j) \in [N_{\mathrm{train}}]$.
	\item[Population] A set of individuals. 
	\item[Fitness] For a given permutation of training example indices, fitness is defined as the average cross entropy loss over training examples when evaluated as in \reffig{fig:schematic}.  The cross entropy loss is computed over the set of possible labels for classifications tasks, and over all tokens in the vocabulary for knowledge base completion tasks.
			
	Note that during search, a training example may occur both in the prompt and as well as the test example. This is generally not a problem as we are not finetuning the model and do not run the risk of learning to copy. When also training the separator token (\refsec{subsec:method_separator}), we ensure that the test example doesn't occur in the prompt by dropping it from the prompt if required.
	\item[Selection] For selection, we use elitism, i.e., at each generation of individuals, we retain a certain percentage (elite ratio) of top performing individuals without any modifications. The rest of the population is created through crossover and mutation over a percentage (selection size) of top performing individuals.
	\item[Crossover] We perform a single point crossover, while ensuring that the resulting individuals contain unique indices. Given two parents $c_1$ and $c_2$, first a random number $j$ is sampled in the range $[k]$, the length of the individuals, to use as the crossover point. We define an operator $\mathrm{First}_s(v,v')$ which selects the first $s$ elements in vector $v$ which do not occur in vector $v'$. Similarly, $\mathrm{Last}_s(v,v')$ picks the last $s$ elements in $v$ which do not occur in vector $v'$. Denoting the subvector $c(i)c(i+1)...c(j)$ by $c^{i:j}$, four new individuals are then created:
	\begin{align*}
		d_1 &= (c_1^{1:j} \mathrm{Last}_{k-j}(c_2,c_1^{1:j})) \\
		d_2 &= (c_2^{1:j} \mathrm{Last}_{k-j}(c_1,c_2^{1:j})) \\
		d_3 &= (\mathrm{First}_{j}(c_2,c_1^{j+1:k}) c_1^{j+1:k}) \\
		d_4 &= (\mathrm{First}_{j}(c_1,c_2^{j+1:k}) c_2^{j+1:k})
	\end{align*}
This modification over a straightforward crossover ensures that the resulting individuals contain unique indices.
	\item[Mutation] We perform mutation on an input candidate by changing each position with a mutation probability $p_m$. When changed, an index is replaced by a random choice from the other training examples. If the new index is already present in the input candidate, the value at that index is swapped with the selected index.
\end{description}

The Genetic algorithm is run for $N_\mathrm{epochs}$ (see \refsec{subsec:background_genetic} for the training flow). A validation set of the same size as the train set was used to select from the best performing individuals in each epoch.

\subsection{Separator Token Learning}
\label{subsec:method_separator}

In addition to the search over permutations of training examples as described in the previous section, we optionally learn a separator token to concatenate the examples (see \reffig{fig:schematic}).

We initialize a token embedding parameter with the token embedding of the newline character. At the end of each epoch of the genetic algorithm, we use gradient updates to estimate the token embedding. The training set is created using the individuals (prompts) in the population in the current generation, and replacing the answer of the final example with the mask token. Gradient updates are then done by requiring the model to predict the correct answer.

\section{Experiments}
\label{sec:expts}

In this section, we aim to answer the following questions:

\begin{description} 
	\item[Q1] How does \system{} compare with existing approaches on task performance? (\refsec{subsec:expts_overall})
	\item[Q2] How do the components of \system{}, namely genetic algorithm and separator token learning affect task performance? (\refsec{subsec:ablation})
	\item[Q3] What aspects of \system{} are important for getting good performance? (\refsec{subsec:expts_analyze})
\end{description}

The experimental setup is described in \refsec{subsec:expts_setup}, and the datasets are described in \refsec{subsec:expts_datasets}.

\subsection{Datasets}
\label{subsec:expts_datasets}
\paragraph{Sentiment Classification:} We use SST2 \citep{socher2013recursive}, a binary sentiment classification task. The training data contains 67350 training, 873 validation and 1822 test examples. 

\paragraph{NLI:} We use label-balanced 2-class NLI dataset created by \citet{shin-etal-2020-autoprompt} using the SICK-E dataset \citep{marelli-etal-2014-sick}. This dataset has 1289 training, 143 validation and 1427 test examples. 

\paragraph{Fact Retrieval:} We use the train, validation, and test splits created by \citet{shin-etal-2020-autoprompt} (referred to as `original' in the paper) for 41 relations. For our experiments, we use the manual prompts created by \citet{petroni-etal-2019-language}. Please see \refapp{sec:fr_details} for relation wise prompts and training statistics.

\begin{table}[t]
\small
\centering
\begin{tabular}{|c|l|c|}
\hline
Training & Prompt Type & P@1 \\ \hline \hline
Manual & LAMA \citep{petroni-etal-2019-language} & 31.1\textasciicircum \\ \hline \hline
\multirow{3}{*}{\begin{tabular}[c]{@{}c@{}}Full\\ supervision\\ \end{tabular}} & LPAQA(top1) \citep{jiang-etal-2020-know} & 34.1\textasciicircum \\ \cline{2-3} 
 & Autoprompt \citep{shin-etal-2020-autoprompt} & 43.3\textasciicircum \\ \cline{2-3} 
 & PERO & 46.6 \\ \hline \hline
\multirow{2}{*}{\begin{tabular}[c]{@{}c@{}}10\\  examples\end{tabular}} & Autoprompt \citep{shin-etal-2020-autoprompt} & 18.9 \\ \cline{2-3} 
 & PERO & 40.3 \\ \hline
\end{tabular}
\caption{\label{tbl:fr_summary} \textit{Summary of fact retrieval experiments:} Precision @1 results on test sets are presented. \textasciicircum indicates replicated numbers. \system{} improves over the baselines when using all training data (up to 1000 examples) and provides significant gains when using limited training data. This indicates that \system{} is  more efficient at eliciting knowledge from language models. Please see \refsec{subsec:expts_overall} for details and \refapp{sec:fr_details} for more detailed results.} 
\end{table}

\subsection{Experimental Setup}
\label{subsec:expts_setup}

\paragraph{Number of training examples:} For most of our experiments, we limit to a total of 10 training examples. We chose this number as prior work \citep{shin-etal-2020-autoprompt} faced difficulty in enabling predictions using only 10 training examples, usually performing close to random prediction. We create 5 sets of size 10, chosen successively from the first 50 training examples, and report on average task performance. Although our focus is few-shot learning in the low data regime, we also present results with more examples (the first 100 and the first 1000 examples) for reference. For model selection, we use a label-balanced validation set (chosen from the beginning of the corresponding validation set) of the same size as the training data.  In all cases, and irrespective of the number of training examples, we keep the prompt size fixed to 10 examples.

\paragraph{Pretrained LM:} We use RoBERTa-large \citep{liu2019roberta} for all our experiments except for the fact retrieval task  where we use the bert-large-cased model \citep{devlin-etal-2019-bert} as this model has been shown to work better for the task \citep{shin-etal-2020-autoprompt}. RoBERTa-large has 24 layers, with 16 attention heads and a hidden size of 1024 (355M parameters). Bert-large-cased uses the same architecture as RoBERTa-large. We use the implementation of transformer architectures provided by \citet{wolf2019huggingface}. We use the \textit{\textless /s\textgreater} token as the default separator token. When learning a new separator token, we initialize the token embedding by the token embedding of \textit{\textless /s\textgreater} token, and finetune the embedding as discussed in \refsec{subsec:method_separator}.

\paragraph{Genetic algorithm:} We run the genetic algorithm for 100 epochs for classification tasks and 30 epochs for fact retrieval tasks. The population size was fixed to 100 and the mutation probability  was set to 0.1. Elite ratio was set to 0.1, while the selection size was fixed to 25. When training a separator token embedding, the maximum number of training epochs for learning the embedding was set to 10 for classification tasks and 5 for fact retrieval tasks. Gradient updates were performed using the AdamW optimizer \citep{loshchilov2018decoupled} with a learning rate of $1e-4$.

\paragraph{Baselines:} We use Autoprompt \cite{shin-etal-2020-autoprompt} and the traditional finetuning approach as few-shot baselines. Please see \refapp{sec:experimental_setup} for hyperparameter details.

\begin{table*}[t]
\small
\centering
\begin{tabular}{|C{2.2cm}|L{2.2cm}|C{1.8cm}|C{1.8cm}|C{1.8cm}|}
\hline
\multicolumn{2}{|l|}{Number of training examples} & \multicolumn{1}{c|}{10} & \multicolumn{1}{c|}{100} & \multicolumn{1}{c|}{1000} \\ \hline \hline
\multirow{3}{*}{\begin{tabular}[c]{@{}c@{}}Sentiment\\ Classification\end{tabular}} & Finetune & 52.5 (2.36) & 90.2 & 93.1 \\ \cline{2-5} 
 & Autoprompt & 52.3 (2.60) & 73.5 & 75.1 \\ \cline{2-5} 
 & PERO & \textbf{91.2} (1.83) & \textbf{93.8} & \textbf{94.2} \\ \hline \hline
\multirow{3}{*}{NLI} & Finetune & 57.4 (10.65) & \textbf{96.1} & \textbf{98.6} \\ \cline{2-5} 
 & Autoprompt & 58.6 (9.08) & 76.2 & 86.5 \\ \cline{2-5} 
 & PERO & \textbf{81.3} (4.99) & 78.5 & 83.2 \\ \hline
\end{tabular}
\caption{\label{tbl:overall_classification} \textit{Summary of classification tasks:} Test set label accuracies for \system{} are presented for the tasks of sentiment classification and NLI. When using 10 examples, we also report the standard deviation across training splits. For both tasks, \system{} provides significant gains over both Autoprompt and the traditional finetuning approach, when using 10 examples. For reference, we also present results when using more training examples. For both tasks, \system{} is competitive with Autoprompt  with increasing data. Overall, the results indicate that \system{} is capable of learning to generalize with a handful of examples, in contrast to existing approaches. Please see \refsec{subsec:expts_overall} for details. Please see \refapp{sec:additional_results_sentiment} for additional comparison between Autoprompt and \system{} using 10 examples across 100 training splits. }
\end{table*}

\subsection{Overall Results}
\label{subsec:expts_overall}

\begin{table}[tb]
\small
\centering
\begin{tabular}{|l|c|c|}
\hline
 & \begin{tabular}[c]{@{}c@{}}Sentiment\\ Classification\end{tabular} & NLI \\ \hline
PERO & 91.2 & 81.3 \\ \hline
PERO-Sep learning & 89.3 & 77.5 \\ \hline
\end{tabular}
  \caption{\label{tbl:ablation} \textit{Impact of separator token learning Step:} Average test set label accuracies with and without the separator token learning are presented for training sets of size 10. The results indicate that the genetic algorithm step alone provides a strong baseline, while the separator token learning step provides further gains. Please see \refsec{subsec:ablation} for details.}
\end{table}

In this section, we present the few-shot learning capability of \system{}. For reference, we also report results when using more data.

We present fact retrieval results (Precision@1 scores) in \reftbl{tbl:fr_summary}. Relation wise results are provided in \refapp{sec:fr_details}. When using all training data, \system{} improves the overall P@1, and is competitive or outperforms Autoprompt on all relations. When using only 10 training examples, \system{} provides significant gains over Autoprompt on all relations.  Overall, we show through \system{} that simple manual prompts can be combined in relatively straightforward ways to create stronger probes while still being interpretable.\footnote{Autoprompt's learned tokens are sometimes relevant to the task but generally hard to interpret.}

We present the label accuracies of \system{} for sentiment classification and NLI in \reftbl{tbl:overall_classification}. In each case, \system{} is able to generalize well when using only 10 examples, while existing approaches perform close to random guess ($~50\%$).
When using more data, \system{} is competitive with Autoprompt for both tasks, while finetuning does better than \system{} for NLI with larger training sizes. Overall, \system{} provides an efficient approach to few-shot learning with pretrained language models.

\paragraph{Comparison when using larger training sizes:}  The results in \reftbl{tbl:overall_classification}  also suggest the use of finetuning when more data is available and the use of \system{} when there isn't enough data for finetuning to generalize well. The relatively low performance of \system{} with more data, especially for the NLI task, could be due to the much larger search space when using more training data. Since we keep the prompt size fixed to 10 examples, the search space is $10!$ for 10 training examples and $1000!/990!$ when using 1000 examples. While a better search strategy could potentially improve \system{}'s performance when using more data, we leave that as an interesting future work. Note, however, that the search space complexity is determined by the number of training examples irrespective of their labels. For example, \system{} improves over the baselines on the fact retrieval task (\reffig{tbl:fr_summary}), despite a much larger number of labels.

For reference, we provide the label accuracies when using all available training data when using \system{}, Autoprompt and finetuning respectively: $95.0$, $91.4$ and $96.7$ for sentiment classification, and $79.5$, $87.3$ and $99.1$ for NLI. When compared to the traditional fully supervised finetuning approach, \system{} performs within $94.3\%$ while using only $0.015\%$ of the training data for sentiment classification, and within $82.1\%$ while using only $0.77\%$ of the training data for NLI.

\subsection{Ablation on \system{}'s components}
\label{subsec:ablation}

In this section, we present an ablation study to understand the role that the two components of \system{}, namely genetic algorithm and separator token learning steps play. We present the label accuracies for sentiment classification and NLI with and without the separator token learning step (indicated as PERO-sep learning) in \reftbl{tbl:ablation}. The results indicate that the permutation search using the genetic algorithm step provides large gains by itself, while the separator token learning potentially improves it.

\begin{table}[t]
\small
\centering
\begin{tabular}{|l|c|c|}
\hline
 & \begin{tabular}[c]{@{}c@{}}Sentiment\\ Classification\end{tabular} & NLI \\ \hline
Default fitness & 89.3 & 77.5 \\ \hline
Inverse fitness & 78.3 & 58.8 \\ \hline
\end{tabular}
\caption{\label{tbl:good_bad_permutations} \textit{Searching for bad permutations:} We use the genetic algorithm described in \refsec{sec:method} to search for bad permutations, by inverting the definition of fitness and report on test accuracy. Average results when using training sizes of 10 examples are presented and contrasted with search using default fitness. The results indicate that good permutations found by \system{} are not trivial and that choosing the right permutation is necessary for generalization. Please see \refsec{subsubsec:bad_solutions} for details.}
\end{table}

\subsection{Analyzing Learned Permutations}
\label{subsec:expts_analyze}   

\subsubsection{Do bad solutions exist which \system{} learns to avoid?}
\label{subsubsec:bad_solutions}
With the same search strategy as discussed in \refsec{sec:method}, we search for potentially bad permutations, by inverting the definition of fitness. For this experiment, to focus on the role of permutations, we do not train the separator token for this experiment. We present the average test set accuracies across training splits for the best and the worst permutations in \reftbl{tbl:good_bad_permutations}. Additionally, we also evaluate 100 random permutations for each training split. The mean (and standard deviation) test accuracy across training splits and random permutations was 85.6(9.08) for sentiment classification and 67.9(8.99) for NLI.

The results indicate that \system{}'s learned permutations provide significant gains over other permutations constructed using the same examples. Selecting the right permutation, therefore, is important for generalization. 

\begin{table}[tb]
\small
\centering
\begin{tabular}{|c|c|c|c|}
\hline
\multicolumn{4}{|c|}{Accuracies on SST2 dev set using only 2 examples} \\ \hline
Label sequence & PERO-10 & Best-2 & Worst-2 \\ \hline
\texttt{----++++--} & 91.4 & 91.5 & 81.2 \\ \hline
\texttt{-++----+--} & 81.1 & 89.5 & 66.5 \\ \hline
\texttt{++++---+++} & 81.4 & 85.5 & 53.9 \\ \hline
\texttt{---+++++--} & 91.5 & 87.4 & 69.3 \\ \hline
\texttt{----++++--} & 91.4 & 90.8 & 74.9 \\ \hline
\end{tabular}
\caption{\label{tbl:using_only_two} \textit{Evaluating learned label patterns with one example per label:} Best and worst accuracies obtained when using only two examples (one unique example per label) are compared with the accuracy of \system{} with 10 distinct examples for the same label pattern (denoted by PERO-10). The best accuracy possible with two distinct examples is competitive with PERO-10. The results indicate that learned label patterns are useful for generalization, along with the choice of selected examples. Please see \refsec{subsubsec:how_many} for details.}
\end{table}

\begin{table}[tb]
\small
\centering
\begin{tabular}{|c|c|c|c|}
\hline
\multicolumn{2}{|c|}{} & \begin{tabular}[c]{@{}c@{}}Sentiment\\ Classification\end{tabular} & NLI \\ \hline \hline
\multirow{2}{*}{\begin{tabular}[c]{@{}c@{}}Proposed\\ 1-shot\end{tabular}} & Worst & 56.2 & 56.3 \\ \cline{2-4} 
 & Best & 90.6 & 84.5 \\ \hline \hline
\multirow{3}{*}{\begin{tabular}[c]{@{}c@{}}10\\ examples\end{tabular}} & Finetune & 52.5 & 57.4 \\ \cline{2-4} 
 & Autoprompt & 52.3 & 58.6 \\ \cline{2-4} 
 & PERO & 91.2 & 81.3 \\ \hline
\end{tabular}
\caption{\label{tbl:one_shot} \textit{One-shot learning:} Best and worst test set label accuracies with one-shot learning using training example pairs obtained from the first 10 training examples are presented. The best possible accuracies with the proposed one-shot learning approach is competitive with \system{} using 10 examples, while improving over finetuning and Autoprompt using 10 examples. Please \refsec{subsubsec:one_shot} for details.}
\end{table}

\begin{table*}[tb]
\small
\centering
\begin{tabular}{|c|l|}
\hline
\multicolumn{2}{|c|}{SST-2} \\ \hline
\begin{tabular}[c]{@{}c@{}}Best\\ (Acc: 90.6 )\end{tabular} & \begin{tabular}[c]{@{}l@{}}-ve sentiment: on the worst revenge-of-the-nerds clichés the filmmakers could dredge up\\ +ve sentiment: demonstrates that the director of such hollywood blockbusters as patriot games\\   can still turn out a small , personal film with an emotional wallop .\end{tabular} \\ \hline
\begin{tabular}[c]{@{}c@{}}Worst\\ (Acc: 56.2)\end{tabular} & \begin{tabular}[c]{@{}l@{}}-ve sentiment: remains utterly satisfied to remain the same throughout\\ +ve sentiment: of saucy\end{tabular} \\ \hline
\end{tabular}
\caption{\label{tbl:one_shot_examples} \textit{Example training pairs for one-shot learning } corresponding to the best and worst test set accuracies for sentiment classification. Please see \refsec{subsubsec:one_shot} for details.}
\end{table*}

\subsubsection{How many examples does \system{} need for good performance?}
\label{subsubsec:how_many}
One could see a permutation learned by \system{} as a combination of a label pattern (the sequence of labels corresponding to the sequence of examples) and particular examples of the respective labels. To understand the importance of the learned label pattern, we search for pairs of examples, one example for each label\footnote{Learning from one example per class is usually referred to as one-shot learning.}, and repeat them using the learned label pattern. The examples are selected from within the set of examples in the learned permutation. We present the accuracy with the original permutation and the best and worst accuracies when using only two training examples in \reftbl{tbl:using_only_two}. Remarkably, two examples alone, when selected well, can go a long way towards good performance.

Additionally, using the learned label pattern provides at least a 10 point improvement in accuracy when compared with a sequence without repetitions (details omitted). This indicates a potential recipe for one-shot learning which we discuss next.

\subsubsection{Can insights gained from this work lead to one-shot learning recipes?}
\label{subsubsec:one_shot}
To answer this question, we provide an example one-shot learning (one training example per class) algorithm which greedily grows a prompt sequence. In contrast to \refsec{sec:method}, we don't use an additional validation set to select a good prompt sequence. We update the definition of fitness to prevent it from being biased towards one class by defining it to be the minimum and not the average of the cross entropy loss over the training examples.   
This is equivalent to minimizing the negative probability of the least probable target label.

Following \refsec{subsubsec:how_many}, we allow an example to be repeated in a prompt sequence. Setting the maximum possible length of the prompt sequence, i.e., number of (potentially repeated) examples in the prompt sequence to $l_\mathrm{max}$, the algorithm then is comprised of the following steps:

\begin{enumerate} 
	\item Initialize an empty prompt, $c=()$ 
	\item Create all possible prompts, $P$, formed by inserting exactly one example to $c$. If we denote the length of $c$ as $l_c$ and the number of labels as $N_\mathrm{labels}$, the size of the set is given by $N_P=(l_c+1)*N_\mathrm{labels}$.
	\item Compute the fitness of prompts in $P$.
	\item Select prompt $c' \in P$ with the best fitness. 
	\item Set $c'=c$ and go to step 2 if $l_c < l_\mathrm{max}$.
\end{enumerate}

We now discuss the results of using this one-shot learning approach over the tasks of sentiment classification and NLI. In each case, we consider the first 10 examples in the training set and create all possible training example pairs for one-shot learning, selecting one example from each class. This leads to 24 training example pairs in each case. We set the max length $l_\mathrm{max}$ to 10, and ensure that the prompt sequence is label-balanced at each step. We summarize the results in \reftbl{tbl:one_shot}. The results indicate that the proposed algorithm is an effective approach to one-shot learning. In \reftbl{tbl:one_shot_examples}, we show the training examples corresponding to the best and worst cases for the task of sentiment classification. While there is indication that more representative examples (such as longer examples) are more informative and thus more useful for one-shot learning, we leave a more thorough analysis as interesting future work.
\section{Conclusion}
\label{sec:conclusion}

In this paper, we propose \system{}, a promising approach for few-shot learning, where we formulate learning as search over permutations of training examples, and optionally a separator token. We show the effectiveness of \system{} for few-shot learning on the tasks of sentiment classification, NLI and fact retrieval tasks. We demonstrate that \system{} provides an interpretable and a more accurate way to probe the knowledge contained in pretrained language models. Our analysis of the learned prompts reveals novel insights and cues for further research on few-shot learning, including one-shot learning.

\section*{Acknowledgments}
We thank the anonymous reviewers for their constructive comments. This work is supported by the Ministry of Human Resource Development (Government of India).

\bibliographystyle{styles/acl_natbib}
\bibliography{pero}

\begin{thebibliography}{19}
\expandafter\ifx\csname natexlab\endcsname\relax\def\natexlab#1{#1}\fi

\bibitem[{Brown et~al.(2020)Brown, Mann, Ryder, Subbiah, Kaplan, Dhariwal,
  Neelakantan, Shyam, Sastry, Askell, Agarwal, Herbert-Voss, Krueger, Henighan,
  Child, Ramesh, Ziegler, Wu, Winter, Hesse, Chen, Sigler, Litwin, Gray, Chess,
  Clark, Berner, McCandlish, Radford, Sutskever, and
  Amodei}]{brown2020language}
Tom Brown, Benjamin Mann, Nick Ryder, Melanie Subbiah, Jared~D Kaplan, Prafulla
  Dhariwal, Arvind Neelakantan, Pranav Shyam, Girish Sastry, Amanda Askell,
  Sandhini Agarwal, Ariel Herbert-Voss, Gretchen Krueger, Tom Henighan, Rewon
  Child, Aditya Ramesh, Daniel Ziegler, Jeffrey Wu, Clemens Winter, Chris
  Hesse, Mark Chen, Eric Sigler, Mateusz Litwin, Scott Gray, Benjamin Chess,
  Jack Clark, Christopher Berner, Sam McCandlish, Alec Radford, Ilya Sutskever,
  and Dario Amodei. 2020.
\newblock \href
  {https://proceedings.neurips.cc/paper/2020/file/1457c0d6bfcb4967418bfb8ac142f64a-Paper.pdf}
  {Language models are few-shot learners}.
\newblock In \emph{Advances in Neural Information Processing Systems},
  volume~33, pages 1877--1901. Curran Associates, Inc.

\bibitem[{Devlin et~al.(2019)Devlin, Chang, Lee, and
  Toutanova}]{devlin-etal-2019-bert}
Jacob Devlin, Ming-Wei Chang, Kenton Lee, and Kristina Toutanova. 2019.
\newblock \href {https://doi.org/10.18653/v1/N19-1423} {{BERT}: Pre-training of
  deep bidirectional transformers for language understanding}.
\newblock In \emph{Proceedings of the 2019 Conference of the North {A}merican
  Chapter of the Association for Computational Linguistics: Human Language
  Technologies, Volume 1 (Long and Short Papers)}, pages 4171--4186,
  Minneapolis, Minnesota. Association for Computational Linguistics.

\bibitem[{Hospedales et~al.(2020)Hospedales, Antoniou, Micaelli, and
  Storkey}]{hospedales2020meta}
Timothy Hospedales, Antreas Antoniou, Paul Micaelli, and Amos Storkey. 2020.
\newblock Meta-learning in neural networks: A survey.
\newblock \emph{arXiv preprint arXiv:2004.05439}.

\bibitem[{Jiang et~al.(2020)Jiang, Xu, Araki, and
  Neubig}]{jiang-etal-2020-know}
Zhengbao Jiang, Frank~F. Xu, Jun Araki, and Graham Neubig. 2020.
\newblock \href {https://doi.org/10.1162/tacl_a_00324} {How can we know what
  language models know?}
\newblock \emph{Transactions of the Association for Computational Linguistics},
  8:423--438.

\bibitem[{Liu et~al.(2019)Liu, Ott, Goyal, Du, Joshi, Chen, Levy, Lewis,
  Zettlemoyer, and Stoyanov}]{liu2019roberta}
Yinhan Liu, Myle Ott, Naman Goyal, Jingfei Du, Mandar Joshi, Danqi Chen, Omer
  Levy, Mike Lewis, Luke Zettlemoyer, and Veselin Stoyanov. 2019.
\newblock {RoBERTa}: A robustly optimized {BERT} pretraining approach.
\newblock \emph{arXiv preprint arXiv:1907.11692}.

\bibitem[{Loshchilov and Hutter(2018)}]{loshchilov2018decoupled}
Ilya Loshchilov and Frank Hutter. 2018.
\newblock Decoupled weight decay regularization.
\newblock In \emph{International Conference on Learning Representations}.

\bibitem[{Marelli et~al.(2014)Marelli, Menini, Baroni, Bentivogli, Bernardi,
  and Zamparelli}]{marelli-etal-2014-sick}
Marco Marelli, Stefano Menini, Marco Baroni, Luisa Bentivogli, Raffaella
  Bernardi, and Roberto Zamparelli. 2014.
\newblock \href
  {http://www.lrec-conf.org/proceedings/lrec2014/pdf/363_Paper.pdf} {A {SICK}
  cure for the evaluation of compositional distributional semantic models}.
\newblock In \emph{Proceedings of the Ninth International Conference on
  Language Resources and Evaluation ({LREC}'14)}, pages 216--223, Reykjavik,
  Iceland. European Language Resources Association (ELRA).

\bibitem[{Mitchell(1998)}]{mitchell1998introduction}
Melanie Mitchell. 1998.
\newblock \emph{An introduction to genetic algorithms}.
\newblock MIT press.

\bibitem[{Mosbach et~al.(2020)Mosbach, Andriushchenko, and
  Klakow}]{mosbach2020stability}
Marius Mosbach, Maksym Andriushchenko, and Dietrich Klakow. 2020.
\newblock On the stability of fine-tuning bert: Misconceptions, explanations,
  and strong baselines.
\newblock In \emph{International Conference on Learning Representations}.

\bibitem[{Petroni et~al.(2019)Petroni, Rockt{\"a}schel, Riedel, Lewis, Bakhtin,
  Wu, and Miller}]{petroni-etal-2019-language}
Fabio Petroni, Tim Rockt{\"a}schel, Sebastian Riedel, Patrick Lewis, Anton
  Bakhtin, Yuxiang Wu, and Alexander Miller. 2019.
\newblock \href {https://doi.org/10.18653/v1/D19-1250} {Language models as
  knowledge bases?}
\newblock In \emph{Proceedings of the 2019 Conference on Empirical Methods in
  Natural Language Processing and the 9th International Joint Conference on
  Natural Language Processing (EMNLP-IJCNLP)}, pages 2463--2473, Hong Kong,
  China. Association for Computational Linguistics.

\bibitem[{Radford et~al.(2018)Radford, Narasimhan, Salimans, and
  Sutskever}]{radford2018improving}
Alec Radford, Karthik Narasimhan, Tim Salimans, and Ilya Sutskever. 2018.
\newblock Improving language understanding by generative pre-training.

\bibitem[{Radford et~al.(2019)Radford, Wu, Child, Luan, Amodei, and
  Sutskever}]{radford2019language}
Alec Radford, Jeff Wu, Rewon Child, David Luan, Dario Amodei, and Ilya
  Sutskever. 2019.
\newblock Language models are unsupervised multitask learners.

\bibitem[{Raffel et~al.(2020)Raffel, Shazeer, Roberts, Lee, Narang, Matena,
  Zhou, Li, and Liu}]{raffel2019exploring}
Colin Raffel, Noam Shazeer, Adam Roberts, Katherine Lee, Sharan Narang, Michael
  Matena, Yanqi Zhou, Wei Li, and Peter~J. Liu. 2020.
\newblock \href {http://jmlr.org/papers/v21/20-074.html} {Exploring the limits
  of transfer learning with a unified text-to-text transformer}.
\newblock \emph{Journal of Machine Learning Research}, 21(140):1--67.

\bibitem[{Shin et~al.(2020)Shin, Razeghi, Logan~IV, Wallace, and
  Singh}]{shin-etal-2020-autoprompt}
Taylor Shin, Yasaman Razeghi, Robert~L. Logan~IV, Eric Wallace, and Sameer
  Singh. 2020.
\newblock \href {https://doi.org/10.18653/v1/2020.emnlp-main.346}
  {{A}uto{P}rompt: {E}liciting {K}nowledge from {L}anguage {M}odels with
  {A}utomatically {G}enerated {P}rompts}.
\newblock In \emph{Proceedings of the 2020 Conference on Empirical Methods in
  Natural Language Processing (EMNLP)}, pages 4222--4235, Online. Association
  for Computational Linguistics.

\bibitem[{Socher et~al.(2013)Socher, Perelygin, Wu, Chuang, Manning, Ng, and
  Potts}]{socher2013recursive}
Richard Socher, Alex Perelygin, Jean Wu, Jason Chuang, Christopher~D. Manning,
  Andrew Ng, and Christopher Potts. 2013.
\newblock \href {https://www.aclweb.org/anthology/D13-1170} {Recursive deep
  models for semantic compositionality over a sentiment treebank}.
\newblock In \emph{Proceedings of the 2013 Conference on Empirical Methods in
  Natural Language Processing}, pages 1631--1642, Seattle, Washington, USA.
  Association for Computational Linguistics.

\bibitem[{Vaswani et~al.(2017)Vaswani, Shazeer, Parmar, Uszkoreit, Jones,
  Gomez, Kaiser, and Polosukhin}]{vaswani2017attention}
Ashish Vaswani, Noam Shazeer, Niki Parmar, Jakob Uszkoreit, Llion Jones,
  Aidan~N Gomez, \L~ukasz Kaiser, and Illia Polosukhin. 2017.
\newblock \href
  {https://proceedings.neurips.cc/paper/2017/file/3f5ee243547dee91fbd053c1c4a845aa-Paper.pdf}
  {Attention is all you need}.
\newblock In \emph{Advances in Neural Information Processing Systems},
  volume~30. Curran Associates, Inc.

\bibitem[{Wang et~al.(2019)Wang, Pruksachatkun, Nangia, Singh, Michael, Hill,
  Levy, and Bowman}]{wang2019superglue}
Alex Wang, Yada Pruksachatkun, Nikita Nangia, Amanpreet Singh, Julian Michael,
  Felix Hill, Omer Levy, and Samuel Bowman. 2019.
\newblock \href
  {https://proceedings.neurips.cc/paper/2019/file/4496bf24afe7fab6f046bf4923da8de6-Paper.pdf}
  {{SuperGLUE}: A stickier benchmark for general-purpose language understanding
  systems}.
\newblock In \emph{Advances in Neural Information Processing Systems},
  volume~32. Curran Associates, Inc.

\bibitem[{Wang et~al.(2018)Wang, Singh, Michael, Hill, Levy, and
  Bowman}]{wang2018glue}
Alex Wang, Amanpreet Singh, Julian Michael, Felix Hill, Omer Levy, and Samuel~R
  Bowman. 2018.
\newblock {GLUE}: A multi-task benchmark and analysis platform for natural
  language understanding.
\newblock In \emph{International Conference on Learning Representations}.

\bibitem[{Wolf et~al.(2020)Wolf, Debut, Sanh, Chaumond, Delangue, Moi, Cistac,
  Rault, Louf, Funtowicz, Davison, Shleifer, von Platen, Ma, Jernite, Plu, Xu,
  Le~Scao, Gugger, Drame, Lhoest, and Rush}]{wolf2019huggingface}
Thomas Wolf, Lysandre Debut, Victor Sanh, Julien Chaumond, Clement Delangue,
  Anthony Moi, Pierric Cistac, Tim Rault, Remi Louf, Morgan Funtowicz, Joe
  Davison, Sam Shleifer, Patrick von Platen, Clara Ma, Yacine Jernite, Julien
  Plu, Canwen Xu, Teven Le~Scao, Sylvain Gugger, Mariama Drame, Quentin Lhoest,
  and Alexander Rush. 2020.
\newblock \href {https://doi.org/10.18653/v1/2020.emnlp-demos.6} {Transformers:
  State-of-the-art natural language processing}.
\newblock In \emph{Proceedings of the 2020 Conference on Empirical Methods in
  Natural Language Processing: System Demonstrations}, pages 38--45, Online.
  Association for Computational Linguistics.

\end{thebibliography}

\appendix
\section{Appendix}

\subsection{Experimental Setup}
\label{sec:experimental_setup}
\subsubsection{Autoprompt Experiments}
For Autoprompt experiments, following \citet{shin-etal-2020-autoprompt}, we set the number of trigger tokens to 10, number of label tokens to 3, and candidate set size to 10. Label search was run for 50 iterations and trigger token search was run for 180 iterations. Experiments were conducted on the same splits as \system{}.

\subsubsection{Finetuning Experiments}
For the finetuning experiments, following the recommended settings for small datasets by \citet{mosbach2020stability}, we trained models for 20 epochs, using AdamW \citep{loshchilov2018decoupled}, with learning rate linearly increasing to $2e-5$ in the first $10\%$ epochs and then linearly decreasing to 0. The experiments were conducted on the same splits as \system{}.

\subsubsection{Training Time}
Training time for \system{} was approximately 3 hours for each experiment in the case of classification tasks, and approximately 30 minutes for each experiment of fact retrieval tasks. 

\subsubsection{Computing Infrastructure}
We used Nvidia's GeForce GTX 1080 Ti GPUs for all our models. Each experiment was run on a single GPU.

\subsubsection{Data}
The experiments were done in the evaluation framework of \citet{shin-etal-2020-autoprompt} who provide instructions for downloading the corresponding data splits at \url{https://github.com/ucinlp/autoprompt}.

Here, we provide more details on the classification datasets used. Details on the fact retrieval data are presented in \refsec{sec:fr_details}.

\paragraph{Sentiment Classification:} We used the SST-2 dataset, the binarized version of the sentiment classification dataset created by \citet{socher2013recursive}. The training examples are constructed using movie review excerpts collected from \url{rottentomatoes.com} website, and labels obtained using Amazon Mechanical Turk's crowdsourcing platform. The percentage of examples labeled with positive sentiment in train, validation and test sets are $55.78\%$, $50.92\%$ and $49.64\%$ respectively. The number of examples labeled with positive sentiment in the training sets of size 10 used in the work are 4, 3, 7, 5 and 4. See \refsec{subsec:expts_setup} for selection and other details.

\paragraph{NLI:} We use the label-balanced 2-class NLI dataset created by \citet{shin-etal-2020-autoprompt} using the SICK-E dataset \citep{marelli-etal-2014-sick}. The dataset was created using sentences from the 8K ImageFlickr data set\footnote{\url{http://nlp.cs.illinois.edu/
HockenmaierGroup/data.html}} and the SemEval 2012 STS MSRVideo Description data set\footnote{\url{http://www.cs.york.ac.uk/semeval-2012/
task6/index.php?id=data}}. Labels were obtained using Amazon Mechanical Turk's crowdsourcing platform. The number of examples labeled with entailment relation in the training sets of size 10 used in the work are 4, 3, 5, 6 and 5. See \refsec{subsec:expts_setup} for selection and other details.

\begin{table}[tb]
\small
\centering
\begin{tabular}{|c|l|l|}
\hline
\multicolumn{2}{|l|}{Number of training examples} & \multicolumn{1}{c|}{10} \\ \hline \hline
\multirow{2}{*}{\begin{tabular}[c]{@{}c@{}}Sentiment\\ Classification\end{tabular}}  & Autoprompt & 54.73 (4.72) \\ \cline{2-3} 
 & PERO & \textbf{91.81} (1.87) \\ \hline \hline
\multirow{2}{*}{NLI} 
 & Autoprompt & 62.31 (8.23) \\ \cline{2-3} 
 & PERO & \textbf{78.61} (6.73) \\ \hline
\end{tabular}
\caption{\label{tbl:classification_additional_results} \textit{Additional results on classification tasks:} Test set label accuracies (and standard deviation) for AutoPrompt and \system{} are presented for the tasks of sentiment classification and NLI across 100 training splits of size 10.}
\end{table}

\subsection{Additional Results}
\label{sec:additional_results}
\subsubsection{Sentiment Classification}
\label{sec:additional_results_sentiment}
With the experimental setup described in \refsec{sec:expts}, we performed additional comparison between Autoprompt and \system{} by creating 100 training splits of size 10, chosen successively from the first 1000 training examples in each dataset. We report on the average (and standard deviation) test accuracy with Autoprompt and \system{} in \reftbl{tbl:classification_additional_results}.

\subsubsection{Fact Retrieval}
\label{sec:fr_details}
We present relation wise training details and LAMA \cite{petroni-etal-2019-language} prompts which we used for our experiments along with the detailed relation wise test results in \reftbl{tbl:fact_retrieval_detailed}.
\begin{table*}[tbh!]
\small
\centering
\begin{tabular}{|l|l|l|l|l||l|l|}
\hline
\multirow{2}{*}{Relation, Manual Prompt (LAMA) (\#train)}                                                  & \multicolumn{4}{c|}{Full data set}                                                                                              & \multicolumn{2}{c|}{Size 10 dataset}                                                                            \\ \cline{2-7}
                                                                                                           & LAMA  & LPAQA & \begin{tabular}[c]{@{}l@{}}Auto\\ Prompt\end{tabular} & \begin{tabular}[c]{@{}l@{}}PERO\end{tabular} & \begin{tabular}[c]{@{}l@{}}Auto\\ Prompt\end{tabular} & \begin{tabular}[c]{@{}l@{}}PERO\end{tabular} \\ \hline
	P1001, {[}X{]} is a legal term in {[}Y{]} (1000)                                                           &70.47&72.75&82.45&84.88&71.50&79.00\\\hline
P101, {[}X{]} works in the field of {[}Y{]} (864)                                                          &9.91&5.32&12.79&18.25&3.60&11.32\\\hline
P103, The native language of {[}X{]} is {[}Y{]} (1000)                                                     &72.16&72.16&82.09&81.88&23.40&76.64\\\hline
P106, {[}X{]} is a {[}Y{]} by profession (1000)                                                            &0.63&0&14.72&14.30&0.60&6.78\\\hline
P108, {[}X{]} works for {[}Y{]} (376)                                                                      &6.79&5.74&8.62&8.09&1.80&7.57\\\hline
P127, {[}X{]} is owned by {[}Y{]} (548)                                                                    &34.79&32.46&35.95&47.89&13.90&39.13\\\hline
P1303, {[}X{]} plays {[}Y{]} (1000)                                                                        &7.59&18.02&15.38&23.71&15.10&16.23\\\hline
P131, {[}X{]} is located in {[}Y{]} (1000)                                                                 &23.27&22.81&37.46&39.95&12.00&30.58\\\hline
P136, {[}X{]} plays {[}Y{]} music (1000)                                                                   &0.75&16.76&55.42&55.42&9.30&55.81\\\hline
P1376, {[}X{]} is the capital of {[}Y{]} (310)                                                             &73.93&59.83&40.17&56.84&26.10&55.81\\\hline
P138, {[}X{]} is named after {[}Y{]} (856)                                                                 &61.55&59.69&66.05&72.40&18.20&70.26\\\hline
\begin{tabular}[c]{@{}l@{}}P140, {[}X{]} is affiliated with the {[}Y{]} religion \\ (445)\end{tabular}     &0.63&59.83&75.26&63.00&49.30&61.02\\\hline
P1412, {[}X{]} used to communicate in {[}Y{]} (1000)                                                       &65.02&64.71&71.21&74.20&49.80&74.01\\\hline
P159, The headquarter of {[}X{]} is in {[}Y{]} (1000)                                                      &32.37&35.57&35.47&39.71&10.20&28.67\\\hline
P17, {[}X{]} is located in {[}Y{]} (1000)                                                                  &31.29&35.48&52.15&59.14&17.20&56.56\\\hline
P176, {[}X{]} is produced by {[}Y{]} (1000)                                                                &85.64&81.67&87.78&87.88&55.20&82.46\\\hline
P178, {[}X{]} is developed by {[}Y{]} (560)                                                                &62.84&59.12&66.72&67.23&29.50&52.30\\\hline
P19, {[}X{]} was born in {[}Y{]} (1000)                                                                    &21.08&20.87&19.92&22.56&6.50&17.80\\\hline
P190, {[}X{]} and {[}Y{]} are twin cities (895)                                                            &2.41&1.91&2.31&2.61&1.00&2.63\\\hline
P20, {[}X{]} died in {[}Y{]} (1000)                                                                        &27.91&27.91&31.16&32.53&11.90&30.62\\\hline
\begin{tabular}[c]{@{}l@{}}P264, {[}X{]} is represented by music label {[}Y{]}\\  (1000)\end{tabular}      &9.56&10.26&43.82&38.46&9.90&28.76\\\hline
P27, {[}X{]} is {[}Y{]} citizen (1000)                                                                     &0&41.51&46.69&48.96&25.80&46.63\\\hline
P276, {[}X{]} is located in {[}Y{]} (1000)                                                                 &41.5&41.5&44.11&48.38&20.80&42.50\\\hline
P279, {[}X{]} is a subclass of {[}Y{]} (1000)                                                              &30.74&14.75&54.93&63.28&22.40&51.95\\\hline
P30, {[}X{]} is located in {[}Y{]} (1000)                                                                  &25.44&18.56&70.36&79.69&43.80&73.23\\\hline
P31, {[}X{]} is a {[}Y{]} (1000)                                                                           &36.66&36.66&51.95&53.90&15.40&45.55\\\hline
P36, The capital of {[}X{]} is {[}Y{]} (1000)                                                              &62.16&62.16&60.6&63.44&14.70&63.36\\\hline
P361, {[}X{]} is part of {[}Y{]} (1000)                                                                    &23.61&31.44&17.7&41.09&1.70&7.96\\\hline
P364, The original language of {[}X{]} is {[}Y{]} (1000)                                                   &44.51&43.93&48.48&53.04&16.60&44.02\\\hline
P37, The official language of {[}X{]} is {[}Y{]} (311)                                                     &54.55&56.83&62.63&67.29&13.00&57.12\\\hline
P39, {[}X{]} has the position of {[}Y{]} (1000)                                                            &7.96&16.14&30.72&37.33&23.10&33.50\\\hline
P407, {[}X{]} was written in {[}Y{]} (1000)                                                                &59.18&65.22&68.42&72.63&41.80&66.50\\\hline
P413, {[}X{]} plays in {[}Y{]} position (1000)                                                             &0.53&23.74&41.7&41.70&19.10&41.70\\\hline
P449, {[}X{]} was originally aired on {[}Y{]} (1000)                                                       &20.89&9.08&34.39&35.19&15.90&28.06\\\hline
P463, {[}X{]} is a member of {[}Y{]} (679)                                                                 &67.11&57.33&54.22&65.78&25.10&39.20\\\hline
P47, {[}X{]} shares border with {[}Y{]} (1000)                                                             &13.67&13.34&19.52&15.84&5.30&14.51\\\hline
P495, {[}X{]} was created in {[}Y{]} (1000)                                                                &16.5&32.23&36.63&40.37&9.90&37.51\\\hline
P527, {[}X{]} consists of {[}Y{]} (1000)                                                                   &11.07&10.55&25.61&27.66&3.30&23.77\\\hline
\begin{tabular}[c]{@{}l@{}}P530, {[}X{]} maintains diplomatic relations with \\ {[}Y{]} (927)\end{tabular} &2.81&3.92&3.11&3.41&0.90&2.21\\\hline
P740, {[}X{]} was founded in {[}Y{]} (1000)                                                                &7.59&13.68&13.89&15.71&8.80&9.25\\\hline
P937, {[}X{]} used to work in {[}Y{]} (1000)                                                               &29.77&39.1&38.36&44.23&13.80&39.81\\\hline
\end{tabular}
\caption{\label{tbl:fact_retrieval_detailed} \textit{Fact retrieval Precision @1}. \system{} outperforms or matches Autoprompt for all except one relation when using all training data. With 10 examples, \system{} performs significantly better than the Autoprompt for all relations. Please see Section 5.3 in the main text for details.}
\end{table*}

\subsection{Validation Set Results}
\label{sec:validation_results}
In this section, we provide the validation set results omitted from the main text.

For sentiment classification, \system{}'s accuracy on validation set with 10, 100 and 1000 examples respectively are $91.2\%$, $93.8\%$ and $94.1\%$. For NLI, \system{}'s accuracy on validation set with 10, 100 and 1000 examples respectively are $81.3\%$, $78.5\%$ and $83.2\%$. The validation set accuracy of \system{}-Sep learning which was trained on 10 training examples was $91.2\%$ for sentiment classification and $79.4\%$ for NLI.

For fact retrieval. the average P@1 was $48.95$ when using all training data, and $42.56$ when using only 10 training examples.

\end{document}